\documentclass[11pt,a4paper]{article}
\PassOptionsToPackage{breaklinks}{hyperref}
\usepackage[hyperref]{eacl2021}
\usepackage{times}
\usepackage{latexsym}
\usepackage{enumitem}
\usepackage{algorithm}
\usepackage{algpseudocode}
\usepackage{soul}

\usepackage{array}
\newcolumntype{L}[1]{>{\raggedright}m{#1}}
\newcolumntype{C}[1]{>{\centering}m{#1}}
\newcolumntype{R}[1]{>{\raggedleft}m{#1}}

\usepackage{microtype}

\aclfinalcopy 

\usepackage{listings}
\definecolor{maroon}{rgb}{0.5,0,0}
\lstdefinelanguage{XML}
{
  morestring=[b]",
  morestring=[s]{>}{<},
  morecomment=[s]{<?}{?>},
  keywordstyle=\color{maroon},
  morekeywords={document,sentence}
}
\title{Memorization vs.\ Generalization:\\Quantifying Data Leakage in NLP Performance Evaluation}

\author{\bf Aparna Elangovan$^1$, 
  \bf Jiayuan He$^{1,2}$, and
  \bf Karin Verspoor$^1$\\
  $^1$The University of Melbourne, Australia\\
  $^2$RMIT University, Australia \\
  {\tt aparnae@student.unimelb.edu.au} \\
  {\tt \{estrid.he, karin.verspoor\}@unimelb.edu.au}}

\date{}

\begin{document}
\maketitle
\begin{abstract}
Public datasets  are often  used to evaluate the efficacy and generalizability of state-of-the-art methods for many tasks in natural language processing (NLP). 
However, the presence of overlap between the train and test datasets can lead to inflated results, inadvertently evaluating the model's ability to memorize and interpreting it as the ability to generalize. In addition, such data sets may not provide an effective indicator of the performance of these methods in real world scenarios. 
We identify leakage of training data into test data on several publicly available datasets used to evaluate NLP tasks, including named entity recognition and relation extraction, and study them to assess the impact of that leakage on the model's ability to memorize versus generalize. 
\end{abstract}

\section{Introduction}

Shared tasks that provide publicly available datasets in order to evaluate and compare the performance of different methods on the same task and data are common in NLP.  Held-out test sets are typically provided, enabling assessment of the 
generalizability of different methods to previously unseen data. These datasets have played a key role in driving progress in NLP, by defining focus tasks and by making annotated data available to the broader community,
in particular in specialized domains such as biomedicine where data can be difficult to obtain, and quality data annotations require the detailed work of domain experts. Examples of tasks where benchmark data sets exist include open domain question answering (QA) \citep{RN95, joshi-etal-2017-triviaqa} and biomedical named entity recognition \citep{smith2008overview} . 

In the context of machine learning models, effectiveness is typically determined by  the model's ability to both memorize and generalize \citep{pmlr-v80-chatterjee18a}. A model that has huge capacity to memorize will often work well in real world applications, particularly
where large amounts of training data are available~\cite{daelemans2005memory}. The ability of a model to generalize relates to how well the model performs when it is applied on data that may be different from the data used to train the model, in terms of e.g.\ the distribution of vocabulary or other relevant vocabulary. The ability to memorize, taken to the extreme, can be considered equivalent to an exact match lookup table \citep{pmlr-v80-chatterjee18a} and the ability to generalize captures how well it can deal with degrees of variations from the lookup table.  An effective combination of memorization and generalization  can be achieved where a model 
selectively memorizes only those aspects or features that matter in solving a target objective given an input, allowing it to generalize better and to be less susceptible to noise.  

When there is considerable overlap in the training and test data for a task, models 
that memorize more effectively than they generalize may benefit from the structure of the evaluation data, with their performance inflated relative to models that are more robust in generalization.
However, such models may make poor quality predictions outside of the shared task setting.
The \textit{external validity} of these evaluations can therefore be questioned \cite{ferro2018dagstuhl}.

In this paper, we assess the overlap between  the train and test data in publicly available datasets for Named Entity Recognition (NER), Relation Extraction (REL) and Text Classification (CLS) tasks, including SST2 \citep{SST2}, BioCreative \citep{smith2008overview,biocreativeIII} and AIMed \citep{RN29} datasets, and examine the significant impact of not taking into account this overlap on performance evaluation.

We argue that robustness in generalization to unseen data is a key consideration of the performance of a model, and propose a framework to examine inadvertent leakage of data between data set splits, in order to enable more controlled assessment of the memorization vs.\ generalization characteristics of different methods.

\section{Related work}
The issue of memorization vs.\ generalization has been previously discussed in the context of question answering datasets, where, given only a question, a system must output the best answer it can find in available texts.

\citet{lewis2020question} identify 3 distinct issues for open domain QA evaluation: a) \textit{question memorization} -- recall the answer to a question that the model has seen at training time; b) \textit{answer memorization} -- answer novel questions at test time, where the model has seen the answer during training; and c) \textit{generalization} -- question and answer not seen during training time. They find that 58-71\% of test answers occur in the training data in 3 examined data sets, concluding that the majority of the test data does not assess  answer generalization. They also find that 28-34\% have paraphrased questions in training data, and a majority of questions are duplicates differing only by a few words.

Similarly, \citet{min2020} identified repeating forms in QA test sets as a problem. The work proposed a novel template-based approach to splitting questions into paraphrase groups referred to as ``Templates'' and then controlling train/test data splits to ensure that all questions conforming to a given template appear in one segment of the data only. This was tested on the EMR Clinical Question Answering dataset emrQA \cite{pampari-etal-2018-emrqa} and the Overnight dataset \cite{wang-etal-2015-building}; it was demonstrated that models perform significantly worse on test sets where strict division is enforced. This paraphrase-based splitting methodology was also employed in their recent work on emrQA \cite{rawat-etal-2020-entity}.

\section{Approach}
\begin{algorithm}
\small
  \caption{Compute overlap}\label{alg:overlap}
  \begin{algorithmic}[1]
    \Procedure{Compare}{$testset,trainset$}
     \State $totalscore \gets 0$
     \State $n \gets |testset| $
     \For{\texttt{$test_i$} \textbf{in} \texttt{testset}}
        \State $s\gets$ \textsc{bestmatch}$(test_i, trainset)$
        \State $totalscore\gets totalscore + s$
       \EndFor\label{euclidendwhile}
       \State \textbf{return} $totalscore/n$ \Comment{Average score}
    \EndProcedure
    
    \Procedure{bestmatch}{$test_i,trainset$}
     \State $bestscore \gets 0$
     \For{\texttt{$train_j$} \textbf{in} \texttt{trainset}}
        \State $s\gets$ \textsc{similarity}$(test_i, train_j)$
        \If{ $score > bestscore$ }
            \State $bestscore \gets s$
         \EndIf
       \EndFor\label{bestmatch}
       \State \textbf{return} $bestscore$ 
    \EndProcedure
  \end{algorithmic}

\end{algorithm}

\begin{table*}[ht]
\centering
\small{
\begin{tabular}{p{0.05\linewidth}p{0.09\linewidth}p{0.03\linewidth}p{0.05\linewidth}p{0.62\linewidth}}
\hline
\textbf{ Task }& \textbf{Dataset} & \textbf{Score} & \textbf{Split} & \textbf{Example} \\
\hline
 REL & AIMed (R) & 100.0 & Train & Thus, during PROTEIN1 -mediated suppression of cell proliferation, PROTEIN and PROTEIN2 may be important for coordinating cell-cycle progression, DNA replication and repair of damaged DNA.\\ 

  &  & & Test & Thus, during PROTEIN -mediated suppression of cell proliferation, PROTEIN1 and PROTEIN2 may be important for coordinating cell-cycle progression, DNA replication and repair of damaged DNA.\\

NER & BC2GM  & 100.0 & Train & E2F family members  \\
& & & Test & E2F family members (1-5)  \\

CLS & SST2 & 100.0 & Train & good movie .\\
&  & & Test & it 's still not a good movie.\\

CLS & SST2 & 21.8 & Train & herzog is obviously looking for a moral to his fable , but the notion that a strong , unified showing among germany and eastern european jews might have changed 20th-century history is undermined by ahola 's inadequate performance .\\
 &  & & Test & of the unsung heroes of 20th century \\
\hline

\end{tabular}
}
\caption{\label{tab:overlapexample}Examples of train-test matches and the corresponding unigram similarity score.}
\vspace{-6mm}
\end{table*}

A common practice to create a train and test set is to shuffle data instances in a dataset and generate random splits, without taking into account broader context. However, this can inadvertently lead to data leakage from the train set to test set due to the overlaps between similar train and test instances. 

The type of overlap between train and test dataset depends on the type of the NLP task. Generally speaking, the leakage can occur either in the input texts or the annotated outputs. We define the types of overlaps which may occur in several NLP tasks as follows.
\begin{itemize}[topsep=0pt, partopsep=0pt, itemsep=0pt, parsep=0pt]
\item In text classification (CLS) tasks such as sentiment analysis, overall (document-level) similarity in input texts can result in train/test leakage.
\item In named entity recognition (NER) tasks, leakage from train to test data may occur when 
\begin{itemize}[topsep=0pt, partopsep=0pt, itemsep=0pt, parsep=0pt]
    \item[a)] input sentences or passages are similar
    \item[b)] target entities are similar 
\end{itemize}
\item In relation extraction (REL) tasks, leakage may occur when 
\begin{itemize}[topsep=0pt, partopsep=0pt, itemsep=0pt, parsep=0pt]
    \item[a)] input sentences or passages are similar
    \item[b)] participating entities are similar
\end{itemize}

\end{itemize}
We propose a framework for quantifying train-test overlaps, and conduct experiments to show the impact of train-test overlap on model performances. Next, we discuss the proposed framework in Sec.~\ref{sec:framework} and the experimental settings in Sec.~\ref{sec:exp_setting}. We present our findings including the train-test overlaps in several benchmark datasets in Sec.~\ref{sec:results_overlap} and the impact of data leakage in Sec.~\ref{sec:results_exp}.

\section{Method}

\subsection{Datasets}
We examine overlap in the following datasets:
\begin{itemize}[topsep=0pt, partopsep=0pt, itemsep=0pt, parsep=0pt]

\item \textbf{AIMed} - AIMed dataset \citep{RN29} for protein relation extraction (REL)

\item \textbf{BC2GM} - BioCreative II gene mention dataset  \citep{smith2008overview} for NER task

\item \textbf{ChEMU} - Chemical Reactions from Patents \citep{he2020overview} for recognising names of chemicals, an NER task

\item \textbf{BC3ACT} - Biocreative III protein interaction classification (CLS)  \citep{biocreativeIII} 

\item \textbf{SST2} - Stanford Sentiment Analysis Treebank \citep{SST2} used to classify sentiments (CLS) in Glue \citep{wang-etal-2018-glue}

\end{itemize}

The AIMed dataset does not explicitly provide a test set and 10-fold cross validation is used for evaluation in previous works \citep{hsieh-etal-2017-identifying,Zhang2019}. In this paper, we use two types of splits of AIMed to evaluate the impact of data leakage: AIMed (R) which \textbf{R}andomly splits the dataset into 10 folds; and AIMed (U) which splits the dataset into 10 folds such that the documents within each resultant split are \textbf{U}nique (according to the \textit{document ID}) to other splits  across each split. The \textit{document ID} refers to the source document of a data instance, and data instances from the same source document have the same document ID, see example in Appendix~\ref{sec:append:aimedexample}

\subsection{Similarity measurement}
\label{sec:framework}
The pseudo code for measuring similarity is shown in Algorithm~\ref{alg:overlap}. Given a test instance $test_i$, we compute its similarity with the training set using the training instance that is most similar with $test_i$. We then use the average similarity over all the test instances as an indicator to measure the extent of train/test overlap. The function $similarity(\cdot)$ can be any function for text similarity. In this paper, we use a simple bag-of-words approach to compute text similarity. We represent each train/test instance with a count vector of unigrams/bigrams/trigrams, ignoring stopwords, and compute the similarity using the cosine similarity.

\subsection{Evaluate model performance}
\label{sec:exp_setting}
We assess the impact of data leakage on a machine learning model's performance. We split the test sets of BC2GM, ChEMU, BC2ACT and SST2 into four intervals considering four similarity threshold ranges (in terms of unigrams): [0-0.25),[0.25-0.50), [0.50-0.75), and [0.75-1.0]. For example, the test instances in the first interval are most different from the training set with a similarity less than 0.25. This method allows full control of the similarity of instances within each interval, but results in a different number of instances in each interval. Thus, we consider another scenario where we split the test set into 4 quartiles based on similarity ranking, so that the number of samples remain the same in each quartile but the threshold varies as a result.

We finetune a BERT (base and cased) model \citep{devlin-etal-2019-bert} for each dataset using their own training set and compare the performance of the finetuned BERT model on the four different test intervals and test quartiles.

We compare the performances of AIMed (R) with AIMed (U) using 3 different models---\citet{Zhang2019} convolutional residual network,  \citet{hsieh-etal-2017-identifying} Bi-LSTM, and BioBERT \citep{biobert}. Following previous works, 
we preprocess the dataset and replace all non-participating proteins with  neutral name \textit{PROTEIN}, the participating entity pairs with \textit{PROTEIN1} and \textit{PROTEIN2}, so the model only ever sees the pseudo protein names. 

\section{Results}
\subsection{Similarity in datasets}
\label{sec:results_overlap}
\begin{table}
\centering
\begin{tabular}{llrrr}
\hline 
  \textbf{Dataset} &\textbf{Task} & \textbf{uni} & \textbf{bi} & \textbf{tri}  \\ 
\hline
AIMED (R)  & REL &    \textbf{96.95} &   \textbf{82.29} &    \textbf{73.15} \\
AIMED (U)  & REL &    \textbf{67.14} &   36.07 &    20.77 \\
BC2GM ann & NER &    \textbf{70.77} &   19.55 &     5.41 \\
BC2GM text & NER &    33.19 &   13.12 &     4.20 \\
BC3ACT  & CLS &    26.76 &    6.91 &     1.81 \\
ChEMU ann & NER &    \textbf{84.29} &   30.67 &     6.83 \\
ChEMU text & NER &    \textbf{68.45} &   42.39 &    31.63 \\
SST2  & CLS &    46.06 &   17.38 &     1.39 \\
\hline
\end{tabular}
\caption{\label{tab:cosineoverlap}Train-test similarity  using unigrams (uni), bigrams (bi), trigrams (tri). BC2GM and ChEMU are in BRAT standoff format and their similarities are shown for their text files (``text'') and annotation files (``ann''). Similarities beyond 60.0 are highlighed in \textbf{bold}.}
\vspace{-2mm}
\end{table}

Examples of similar train and test instances are shown in Table~\ref{tab:overlapexample}. The overall results of train-test similarities of all datasets are shown in Table~\ref{tab:cosineoverlap}. 

In the BC2GM dataset, we find that there is ~70\% overlap between gene names in the train and test set. On further  analysis, we find that 2,308 out of 6,331 genes in the test set have exact matches in the train set. 
In the AIMed (R) dataset, we can see that there is over 73\% overlap, even measured in the trigrams, between train and test sets.

\subsection{Model performance and similarity}
\label{sec:results_exp}

We observe drops in F-scores of more than 10 points between AIMed (R) and AIMed (U) across all three models as shown in Table~\ref{tab:aimed-results}. This is in line with the similarity measurement in Table~\ref{tab:cosineoverlap}: the train-test similarity drops significantly from AIMed (R) to AIMed (U) since in AIMed (U) we only allow unique document IDs in different folds. 

\begin{table}
\centering
\begin{tabular}{llrrr}
\hline 
\textbf{Split type} & \textbf{Method} & \textbf{P} & \textbf{R} & \textbf{F1}  \\ 
\hline
O & BiLSTM   & 78.8 & 75.2 & 76.9  \\
O & ConvRes  & 79.0	& 76.8 & 77.6 \\
\hline
\multicolumn{5}{c}{Replicated experiments} \\
\hline
R & BiLSTM     &	74.5 &  69.7 & 71.7 \\
U & BiLSTM  &	\textbf{57.4} &	\textbf{61.7} &	\textbf{58.7} \\
R & ConvRes &   71.1 &	69.2 & 69.9 \\
U & ConvRes 	& \textbf{56.7} & \textbf{ 56.4} & \textbf{56.1} \\
R & BioBERT &	79.8 &	76.7 & 77.9 \\
U & BioBERT    & \textbf{65.8} &	\textbf{63.7} &	\textbf{64.4} \\

\hline
\end{tabular}
\caption{\label{tab:aimed-results} Performances on AIMed (R) and AIMed (U). The split type (O) indicates the original results from the authors.}
\vspace{-6mm}
\end{table}

On the ChEMU NER dataset we observe nearly 10-point drop in F-score (96.7$\rightarrow$85.6) from 4I to 2I as shown in Table~\ref{tab:biobertner}. 

On the BC2GM dataset, we also find that the model performance degrades from 82.4\% to 74.5\% in 2I compared to that in 1I. Surprisingly, F-score for 4I is substantially lower than that of 3I (78.5$\rightarrow$87.1), despite 41 out of the total 47 instances in 4I having 100\% similarity with the train set (full detailed samples shown in Appendix~Table~\ref{tab:append:bc2gmsim}). A further investigation on this shows that \textbf{(a)} the interval 4I only has 0.9\% (47/5000) of test instances; \textbf{(b)} a significant drop in recall (90.6$\rightarrow$77.5) from 3I to 4I is caused by six instances whose input texts have exact matches in the train set (full samples shown in Appendix~Table~\ref{tab:append:bc2gmrecallfail}). This implies that the model doesn't perform well even on the training data for these samples.  Since BC2GM has over 70\% overlap in the target gene mentions (Table~\ref{tab:cosineoverlap}), we also analysed the recall on the annotations that overlap between train and test.  We find that the recall increases  (84.5$\rightarrow$87.8), see Appendix~Table~\ref{taba:bertoverlapoanno}, compared to recall (81.1$\rightarrow$90.6) as a result of input text similarity. Since BERT uses a word sequence-based prediction approach, the relatively high similarity in target annotations does not seem to make much difference compared to similarity in input text. However, if we used  a dictionary-based approach, similarity in  annotations could result in much higher recall compared to similarity in input text.

\begin{table}
\centering

\begin{tabular}{lp{0.2cm}rrrrr}
\hline
     \textbf{D}   &  \textbf{SR} &   \textbf{\%} & \textbf{P} & \textbf{R}  & \textbf{F1} & \textbf{A} \\
\hline
   BC2  &   F &   100.0 &      77.5 &   86.4 &    81.7 &       \\
  BC2  &  1I &    19.8 &      68.8 &   81.1 &   \textbf{ 74.5} &      \\
  BC2  &  2I &    74.1 &      78.3 &   86.9 &    82.4 &       \\
  BC2  &  3I &     5.1 &      83.8 &   90.6 &   \textbf{ 87.1} &      \\
  BC2  &  4I &     0.9 &      79.5 &   77.5 &    78.5 &      \\
  \hline
  ChE  & F & 100.0 & 93.8 & 94.4 & 94.1\\
ChE  & 1I & 0.0 & - & - & - \\
ChE  & 2I & 10.0 & 84.6 & 86.6 & \textbf{85.6}\\
ChE  & 3I & \textbf{60.0} & 93.4 & 94.0 & 93.7 \\
ChE  & 4I & 30.0 & 96.7 & 96.7 & \textbf{96.7} \\
\hline
BC3  &   F &   100.0 &      45.1 &   84.1 &    58.7 &     82.1 \\
 BC3  &  1I &    47.0 &      43.0 &   82.0 &    \textbf{56.4 }&     85.8 \\
 BC3  &  2I &    51.0 &      46.0 &   85.6 &    59.9 &     78.8 \\
 BC3  &  3I &     2.0 &      53.5 &   76.7 &    \textbf{63.0} &     77.5 \\
 BC3  &  4I &     0.0 &       0.0 &    0.0 &     0.0 &      0.0 \\
 \hline
    SST  &   F &   100.0 &      90.4 &   96.7 &    93.4 &     93.2 \\
   SST  &  1I &     1.1 &      60.0 &   75.0 &     \textbf{66.7} &      \textbf{85.0} \\
   SST  &  2I &    66.8 &      91.6 &   96.0 &    93.8 &     93.4 \\
   SST  &  3I &    28.7 &      87.1 &   98.7 &    92.5 &     92.7 \\
   SST  &  4I &     3.5 &      96.9 &   96.9 &    \textbf{96.9} &      \textbf{96.8} \\
   
 \hline
   
\end{tabular}
\caption{\label{tab:biobertner}Performances on various similarity thresholds and the corresponding percentage of test instances within the intervals. Datasets (D): BC2$\rightarrow$BC2GM, ChE$\rightarrow$ChEMU, BC3$\rightarrow$BC3ACT, SST$\rightarrow$SST2. The similarity threshold range (SR) $[0,0.25)=1$I, $[0.25,0.5)=2$I, $[0.5,0.75)=3$I, $[0.75,1]=4$I, $[0,1]=$ F. Accuracy (A) is the official metric for the SST2 dataset according to GLUE benchmark, all others use F1-score (F1) as primary metric.}
 \vspace{-6mm}
\end{table}

The BC3ACT dataset also exhibits the same trend where the F1-score improves (56.4$\rightarrow$63.0) as the similarity increases. However, the accuracy drops from 85.8$\rightarrow$77.5. This is could be because while the train set has 50\% positive classes, the test set has just 17\% with 3 points higher mean similarity in positive samples (details in Appendix~\ref{sec:append:classwise}). 

On SST2, an increase in accuracy (85.0 $\rightarrow$ 96.8) from 1I to 4I is observed apart from a marginal 0.7-point drop  (93.4 $\rightarrow$ 92.7) from 2I to 3I. 

We also split the test sets into four equal-sized quartiles based on the similarity ranking of test instances, 
shown in Table~\ref{tab:simialritysplitquartile}. We observe similar phenomena as in the previous set of experiments for the dataset BC2GM, ChEMU, and BC3ACT. The only exception is for SST2 where the F-score has a relatively small but consistent increase from Q1 to Q3 (92.9$\rightarrow$94.1) but drops to 92.8 in Q4.

\begin{table}
\centering
   \begin{tabular}{L{0.3cm}R{0.4cm}R{0.6cm}R{0.9cm}R{0.5cm}R{0.5cm}R{0.5cm}m{0.5cm}}
    \hline
      \textbf{D}   & \textbf{Q}   & \textbf{Min} &   \textbf{Max} &  \textbf{P} &  \textbf{R} &  \textbf{F1} &  \textbf{A} \\
    \hline
  BC2  & 1 &  0.0 &  26.3 &      69.8 &   82.0 &    \textbf{75.4} &       \\
  BC2  & 2 & 26.3 &  31.6 &      74.5 &   85.9 &    79.8 &       \\
  BC2  & 3 & 31.6 &  38.3 &      78.3 &   86.4 &    82.1 &       \\
  BC2  & 4 & 38.3 & 100.0 &      83.0 &   88.9 &    \textbf{85.9} &       \\
    \hline

    ChE & 1 & 37.9 &  56.7 &   90.8 & 91.8 & \textbf{91.3} &\\
    ChE & 2 & 56.8 &  68.2 &   93.3 & 94.4 & 93.8 & \\
    ChE & 3 & 68.2 &  78.5 &   95.1 & 96.1 & 95.6 & \\
    ChE & 4 & 78.6 & 99.8 &   97.1 & 97.4 & \textbf{97.3} & \\
\hline
 BC3 & 1  &  6.3 &  20.1 &      44.5 &   81.4 &    \textbf{57.6} &     88.8 \\
 BC3 & 2 & 20.1 &  25.7 &      42.3 &   82.5 &    55.9 &     82.7 \\
 BC3 & 3 & 25.7 &  31.9 &      46.5 &   85.3 &    \textbf{60.2} &     79.5 \\
 BC3 & 4 & 31.9 &  75.0 &      46.1 &   85.2 &    59.8 &     77.3 \\
   \hline
   SST  & 1 &  0.0 &  36.5 &      90.8 &   95.2 &    \textbf{92.9} &     \textbf{92.8} \\
   SST & 2 & 36.5 &  43.6 &      91.3 &   96.2 &    93.7 &     93.2 \\
   SST & 3 & 43.6 &  53.5 &      91.2 &   97.3 &    \textbf{94.2} &     \textbf{94.1} \\
   SST & 4 & 53.5 & 100.0 &      88.0 &   98.1 &    92.8 &     92.9 \\
    \hline
    
    \end{tabular}
    \caption{Performances on four different test quartiles, where the number of samples in each quartile (Q) is kept same. The minimum (Min) and the maximum (Max) similarity within each quartile are also reported.}
    \label{tab:simialritysplitquartile}
    \vspace{-6mm}
\end{table}

\section{Discussion}
\subsection{Quantifying similarity}
The bag-of-words based approach to compute cosine similarity has been able to detect simple forms of overlap effectively as shown in Table~\ref{tab:cosineoverlap}. A trend that can be seen is that overlap is more common in tasks that are manual labour intensive, such as named entity recognition and relation extraction compared to text classification. 

However, this approach may detect similarity even when the meanings are different, especially in the case of classification tasks as shown for SST2  in Table~\ref{tab:overlapexample}. 
Semantic Text Similarity (STS) measurement is a challenging task in its own right, with a large body of literature and a number of shared tasks organized to address it \cite{cer-etal-2017-semeval,wang2020medsts,Karimi}.  More sophisticated methods for similarity measurement developed in these contexts could be incorporated into the framework for measuring similarity of data set splits; for simple leakage detection it is arguably adequate.  However, sophisticated methods can also potentially lead to a chicken and egg problem, if we use a machine learning model to compute semantic similarity.

The question of what level of similarity is acceptable is highly data and task-dependent. If the training data has good  volume and variety, the training-test similarity will naturally be higher and so will the acceptable similarity. 

\subsection{Memorization vs.\ Generalization}
We find that the F-scores tend to be higher when the test set input text is similar to the training set as shown in Table~\ref{tab:aimed-results} and~\ref{tab:biobertner}. While this might be apparent, quantifying similarity in the test set helps understand that high scores in the test set could be a result of similarity to the train set, and therefore measuring memorization and not a model's ability to generalize. If a model is trained on sufficient volume and variety of data then it may now matter if it memorizes or generalizes in a real world context, and a model's ability to memorize is not necessarily a disadvantage. However, in the setting of a shared task, we often do not have access to sufficiently large training data sets and hence it is important to consider the test/train similarity when evaluating the models. This implies that in real world scenarios the model may perform poorly when it encounters data not seen during training. 


\section{Conclusion}
 We conclude that quantifying train/test overlap is crucial to assessing real world applicability of machine learning in NLP tasks, given our reliance on annotated data for training and testing in the NLP community. A single metric 
 over a held-out test set is not sufficient to infer generalizablity of a model. Stratification of test sets by similarity enables more robust assessment of memorization vs.\ generalization capabilities of models. 
 Further development of approaches to structured consideration of model performance under different assumptions will improve our understanding of these tradeoffs.

\bibliography{references}
\bibliographystyle{acl_natbib}

\clearpage
\appendix

\appendix
\onecolumn

\section{AIMed document examples}

\label{sec:append:aimedexample}

The following example shows how multiple data instances are extracted from a single document in AIMed dataset. The document with ID ``\texttt{AIMed.d0}'' has several instances including ``\texttt{AIMed.d0.s0}'' and ``\texttt{AIMed.d0.s1}''. These instances thus have the same document id.

{\tiny
\begin{lstlisting}

<corpus source="AIMed">
  <document id="AIMed.d0">
    <sentence id="AIMed.d0.s0" text="Th1/Th2 type cytokines in hepatitis B patients treated with interferon-alpha." seqId="s0">
      <entity id="AIMed.d0.s0.e0" charOffset="60-75" type="protein" text="interferon-alpha" seqId="e0"/>
    </sentence>
    <sentence id="AIMed.d0.s1" text="OBJECTIVE: To investigate the relationship between the expression of Th1/Th2 type cytokines and the effect of interferon-alpha therapy." seqId="s1">
      <entity id="AIMed.d0.s1.e0" charOffset="110-125" type="protein" text="interferon-alpha" seqId="e1"/>
    </sentence>
  </document>
  <document id="AIMed.d1">
    <sentence id="AIMed.d1.s11" text="Involvement of BMP-2 signaling in a cartilage cap in osteochondroma." seqId="s11">
      <entity id="AIMed.d1.s11.e0" charOffset="15-19" type="protein" text="BMP-2" seqId="e15"/>
    </sentence>
  </document>
</corpus>
\end{lstlisting}
}

\section{Classwise similarity for BC3AST}
\label{sec:append:classwise}
The test set has 5090 negative samples compared to 910 positive samples, with 2.96 points higher mean similarity in positive samples.
\begin{table}[ht]
{\small

    \centering
    \begin{tabular}{llrrr}
\hline
 Test label &     &  Unigram &  Bigram &  Trigram \\

\hline
0 & count & 5090.00 & 5090.00 & 5090.00 \\
  & mean & 26.31 & 6.70 & 1.73 \\
  & std & 9.25 & 5.35 & 1.72 \\
  & min & 6.28 & 0.00 & 0.00 \\
  & 25\% & 19.70 & 3.29 & 0.79 \\
  & 50\% & 25.16 & 5.07 & 1.39 \\
  & 75\% & 31.53 & 8.29 & 2.27 \\
  & max & 75.01 & 41.75 & 18.71 \\
1 & count & 910.00 & 910.00 & 910.00 \\
  & mean & 29.27 & 8.09 & 2.26 \\
  & std & 9.36 & 6.00 & 1.73 \\
  & min & 11.14 & 1.52 & 0.00 \\
  & 25\% & 22.69 & 4.51 & 1.17 \\
  & 50\% & 28.31 & 6.25 & 1.88 \\
  & 75\% & 34.32 & 9.38 & 2.84 \\
  & max & 74.01 & 51.20 & 18.97 \\
\hline
\end{tabular}
    \caption{Class-wise similarity for BC3ACT dataset}
    \label{tab:append:classwisebc3ast}
    }
\end{table}

\section{BERT and similarity thresholds }

\label{sec:append:bertoverlap}
Table~\ref{taba:bertoverlap} shows the impact on precision, recall and F-score using different similarity thresholds on the BC2GM test set, which has approximately 6,300 annotations.

We also compare the recall when the target annotations are similar as shown in Table~\ref{taba:bertoverlapoanno}. We only compare unigrams, as the number of tokens in a gene name tends to be small (on average less than 3). 

\begin{table}[ht]
{\small

\centering
\begin{tabular}{ll rrrrr}
\hline 
\textbf{Dataset}   & \textbf{N} & \textbf{SR} &  \% & \textbf{P} & \textbf{R} & \textbf{F}\\ 
\hline
BC2GM & -   & -  & 100  & 77.5 & 86.4 & 81.7 \\
\hline
BC2GM  & 1  & 1I   & 19.8 & 68.8  & 81.1 & 74.5 \\
BC2GM  & 1  & 2I   & 74.1 & 78.2  & 86.9 & 82.3 \\
BC2GM  & l  & 3I   & 5.1 & 83.8  & 90.6 & 87.1  \\
BC2GM  & l  & 4I  & 1.0 & 79.5  & 77.5 & 78.5  \\
\hline
BC2GM  & 2  & 1I  & 91.7 & 76.9  & 86.3 & 81.4  \\
BC2GM  & 2  & 2I  &  7.5 & 82.5  & 88.1 & 85.2  \\
BC2GM  & 2  & 3I  &  0.3 & 1.0  & 1.0 & 1.0  \\
BC2GM & 2  & 4I  &  0.5 & 78.9  & 76.9 & 77.9  \\
\hline
BC2GM  & 3  & 1I  & 98.5 & 77.4  & 86.4 & 81.7  \\
BC2GM  & 3  & 2I  &  0.9 & 85.2  & 88.5 & 86.8  \\
BC2GM  & 3  & 3I  &  0.1 & 50.0  & 100.0 & 66.7  \\
BC2GM  & 3  & 4I  &  0.5 & 80.6  & 76.3 & 78.4  \\
\hline
\end{tabular}
}
\caption{\label{taba:bertoverlap}NER performances of BERT on various similarity threshold range (SR) and the corresponding percentage of instances when the similarity is computed using N-grams (N = 1, 2 and 3) in the input text. The range $[0,0.25)=1$I, $[0.25,0.5)=2$I, $[0.5,0.75)=3$I, $[0.75,1]=4$I, $[0,1]=$ F.}
\end{table}

\begin{table}[ht]
\centering
{\small
\begin{tabular}{lrlrr}
\hline
      Dataset &  N &  SR & \% & Recall \\
\hline
 BC2GM (anno) &     - &   F &   100.0 &   86.4 \\
 BC2GM (anno) &      1 &  1I &    16.7 &   84.5 \\
 BC2GM (anno) &      1 &  2I &     5.6 &   81.8 \\
 BC2GM (anno) &      1 &  3I &    24.7 &   85.6 \\
 BC2GM (anno) &      1 &  4I &    53.0 &   87.8 \\
\hline
\end{tabular}
}
\caption{\label{taba:bertoverlapoanno}NER score on BERT at various similarity threshold range (SR) and the corresponding \% of samples using ngram N = 1 in the output annotated gene mentions. }
\end{table}

Table~\ref{tab:append:bertssteactsim} shows BERT's performance using bi-grams and trigrams on SST2 and BC3AST datasets.

\begin{table}[ht]
{\small
   \centering
    \begin{tabular}{lrlrrrrr}
\hline
       Dataset &  N &  SR & \% & P & R & F1 & A \\
\hline
 
 BC3ACT  &     -  &   F &   100.0 &      45.1 &   84.1 &    58.7 &     82.1 \\
 BC3ACT  &      1 &  1I &    47.0 &      43.0 &   82.0 &    56.4 &     85.8 \\
 BC3ACT  &      1 &  2I &    51.0 &      46.0 &   85.6 &    59.9 &     78.8 \\
 BC3ACT  &      1 &  3I &     2.0 &      53.5 &   76.7 &    63.0 &     77.5 \\
 BC3ACT  &      1 &  4I &     0.0 &       0.0 &    0.0 &     0.0 &      0.0 \\
 BC3ACT  &      2 &  1I &    98.2 &      45.0 &   84.1 &    58.6 &     82.2 \\
 BC3ACT  &      2 &  2I &     1.8 &      48.8 &   83.3 &    61.5 &     76.6 \\
 BC3ACT  &      2 &  3I &     0.0 &     100.0 &  100.0 &   100.0 &    100.0 \\
 BC3ACT  &      2 &  4I &     0.0 &       0.0 &    0.0 &     0.0 &      - \\
 BC3ACT  &      3 &  1I &   100.0 &      45.1 &   84.1 &    58.7 &     82.1 \\
 BC3ACT  &      3 &  2I &     0.0 &       0.0 &    0.0 &     0.0 &      - \\
 BC3ACT  &      3 &  3I &     0.0 &       0.0 &    0.0 &     0.0 &      - \\
 BC3ACT  &      3 &  4I &     0.0 &       0.0 &    0.0 &     0.0 &      - \\
 \hline
   SST2  &     - &   F &   100.0 &      90.4 &   96.7 &    93.4 &     93.2 \\
   SST2  &      1 &  1I &     1.1 &      60.0 &   75.0 &    66.7 &     85.0 \\
   SST2  &      1 &  2I &    66.8 &      91.6 &   96.0 &    93.8 &     93.4 \\
   SST2  &      1 &  3I &    28.7 &      87.1 &   98.7 &    92.5 &     92.7 \\
   SST2  &      1 &  4I &     3.5 &      96.9 &   96.9 &    96.9 &     96.8 \\
   SST2  &      2 &  1I &    64.0 &      88.6 &   96.0 &    92.2 &     92.3 \\
   SST2  &      2 &  2I &    30.8 &      93.1 &   97.3 &    95.1 &     94.8 \\
   SST2  &      2 &  3I &     4.8 &      93.1 &  100.0 &    96.4 &     95.4 \\
   SST2  &      2 &  4I &     0.4 &     100.0 &  100.0 &   100.0 &    100.0 \\
   SST2  &      3 &  1I &    97.6 &      90.5 &   96.6 &    93.4 &     93.3 \\
   SST2  &      3 &  2I &     1.9 &      82.6 &  100.0 &    90.5 &     88.2 \\
   SST2  &      3 &  3I &     0.5 &     100.0 &  100.0 &   100.0 &    100.0 \\
   SST2  &      3 &  4I &     0.0 &       0.0 &    0.0 &     0.0 &      - \\
\hline
\end{tabular}
}
 
    \caption{SST2 and BC3ACT similarity thresholds using ngram N = 1,2 and 3.  The range $[0,0.25)=1I$, $[0.25,0.5)=2I$, $[0.5,0.75)=3I$, $[0.75,1]=4I$, $[0,1]=F$ }
    \label{tab:append:bertssteactsim}
\end{table}

\clearpage

\section{High similarity BC2GM samples}
Table~\ref{tab:append:bc2gmsim} shows the 75\% similarity samples in the BC2GM dataset. The samples that caused the drop in recall are shown in Table~\ref{tab:append:bc2gmrecallfail}.

\begin{table*}[h!]
    \centering
    {\tiny
    \begin{tabular}{p{0.05\linewidth}p{0.45\linewidth}p{0.45\linewidth}}
\hline
   Score &    Test &    Train \\
\hline
76.45 &  Histological and immunophenotypic studies revealed 12 large cell lymphomas (11 B cell and one T cell), two small noncleaved cell lymphomas (B-cell phenotype), and five low grade B-cell lymphomas (two small lymphocytic and three follicular mixed lymphomas). &  The cases included 35 de novo diffuse aggressive lymphomas (DAL; 19 large-cell, 4 mixed-cell, and 12 large-cell immunoblastic), 52 transformed aggressive lymphomas derived from follicular lymphomas (TFL), 42 indolent follicular lymphomas (FL), 14 mantle cell lymphomas (MCL), and 27 small noncleaved cell lymphomas (SNCL). \\
77.46 &  98, 93-98). &  356, 93-98]. \\
81.65 &  Free protein S deficiency in acute ischemic stroke. &  Ischemic stroke due to protein C deficiency. \\
83.41 &  In stage I, histochemistry for copper was positive in 11 out of 21 cases: 6 cases were T+; 1 case R+ and 2 cases O+; 2 cases were T+, R+, O+. &  3 cases \\
86.60 &  STUDY DESIGN: Retrospective review. &  DESIGN: Retrospective study. \\
86.60 &  Non-dialyzable transfer factor &  Dialyzable transfer factor. \\
100.00 &  Recently we have performed a detailed analysis of specific neuronal populations affected by the mutation which shed new light on the role of Krox-20 in the segmentation and on the physiological consequences of its inactivation. &  Recently we have performed a detailed analysis of specific neuronal populations affected by the mutation which shed new light on the role of Krox-20 in the segmentation and on the physiological consequences of its inactivation. \\
100.00 &  Slowly adapting type I mechanoreceptor discharge as a function of dynamic force versus dynamic displacement of glabrous skin of raccoon and squirrel monkey hand. &  Slowly adapting type I mechanoreceptor discharge as a function of dynamic force versus dynamic displacement of glabrous skin of raccoon and squirrel monkey hand. \\
100.00 &  The recruitment of constitutively phosphorylated p185(neu) and the activated mitogenic pathway proteins to this membrane-microfilament interaction site provides a physical model for integrating the assembly of the mitogenic pathway with the transmission of growth factor signal to the cytoskeleton. &  The recruitment of constitutively phosphorylated p185(neu) and the activated mitogenic pathway proteins to this membrane-microfilament interaction site provides a physical model for integrating the assembly of the mitogenic pathway with the transmission of growth factor signal to the cytoskeleton. \\
100.00 &  A heterologous promoter construct containing three repeats of a consensus Sp1 site, cloned upstream of a single copy of the ZII (CREB/ AP1) element from the BZLF1 promoter linked to the beta-globin TATA box, exhibited phorbol ester inducibility. &  A heterologous promoter construct containing three repeats of a consensus Sp1 site, cloned upstream of a single copy of the ZII (CREB/ AP1) element from the BZLF1 promoter linked to the beta-globin TATA box, exhibited phorbol ester inducibility. \\
100.00 &  The reconstituted RNA polymerases containing the mutant alpha subunits were examined for their response to transcription activation by cAMP-CRP and the rrnBP1 UP element. &  The reconstituted RNA polymerases containing the mutant alpha subunits were examined for their response to transcription activation by cAMP-CRP and the rrnBP1 UP element. \\
100.00 &  Analysis of 1 Mb of published sequence from the region of conserved synteny on human chromosome 5q31-q33 identified 45 gene candidates, including 35 expressed genes in the human IL-4 cytokine gene cluster. &  Analysis of 1 Mb of published sequence from the region of conserved synteny on human chromosome 5q31-q33 identified 45 gene candidates, including 35 expressed genes in the human IL-4 cytokine gene cluster. \\
100.00 &  Although RAD17, RAD24 and MEC3 are not required for cell cycle arrest when S phase is inhibited by hydroxyurea (HU), they do contribute to the viability of yeast cells grown in the presence of HU, possibly because they are required for the repair of HU-induced DNA damage. &  Although RAD17, RAD24 and MEC3 are not required for cell cycle arrest when S phase is inhibited by hydroxyurea (HU), they do contribute to the viability of yeast cells grown in the presence of HU, possibly because they are required for the repair of HU-induced DNA damage. \\
100.00 &  The promoter for HMG-CoA synthase contains two binding sites for the sterol regulatory element-binding proteins (SREBPs). &  The promoter for HMG-CoA synthase contains two binding sites for the sterol regulatory element-binding proteins (SREBPs). \\
100.00 &  Coronary vasoconstriction caused by endothelin-1 is enhanced by ischemia-reperfusion and by norepinephrine present in concentrations typically observed after neonatal cardiopulmonary bypass. &  Coronary vasoconstriction caused by endothelin-1 is enhanced by ischemia-reperfusion and by norepinephrine present in concentrations typically observed after neonatal cardiopulmonary bypass. \\
100.00 &  (LH P < 0.05, LH/FSH P < 0.01). &  (LH P < 0.05, LH/FSH P < 0.01). \\
100.00 &  Determinants of recurrent ischaemia and revascularisation procedures after thrombolysis with recombinant tissue plasminogen activator in primary coronary occlusion. &  Determinants of recurrent ischaemia and revascularisation procedures after thrombolysis with recombinant tissue plasminogen activator in primary coronary occlusion. \\
100.00 &  The human SHBG proximal promoter was analyzed by DNase I footprinting, and the functional significance of 6 footprinted regions (FP1-FP6) within the proximal promoter was studied in human HepG2 hepatoblastoma cells. &  The human SHBG proximal promoter was analyzed by DNase I footprinting, and the functional significance of 6 footprinted regions (FP1-FP6) within the proximal promoter was studied in human HepG2 hepatoblastoma cells. \\
100.00 &  Biol. &  Biol. \\
100.00 &  Copyright 1999 Academic Press. &  Copyright 1999 Academic Press. \\
100.00 &  These results demonstrate a specific association of SIV and HIV-2 nef, but not HIV-1 nef, with TCRzeta. &  These results demonstrate a specific association of SIV and HIV-2 nef, but not HIV-1 nef, with TCRzeta. \\
100.00 &  Urease activity, judged as the amount of ammonia production from urea, could be measured at 25 ng per tube (S/N = 1.5) with Jack bean meal urease. &  Urease activity, judged as the amount of ammonia production from urea, could be measured at 25 ng per tube (S/N = 1.5) with Jack bean meal urease. \\
100.00 &  Copyright 1999 Academic Press. &  Copyright 1999 Academic Press. \\
100.00 &  IV. &  IV. \\
100.00 &  Copyright 1998 Academic Press. &  Copyright 1998 Academic Press. \\
100.00 &  IV. &  IV. \\
100.00 &  Biol. &  Biol. \\
100.00 &  Copyright 1999 Academic Press. &  Copyright 1999 Academic Press. \\
100.00 &  Copyright 1998 Academic Press. &  Copyright 1998 Academic Press. \\
100.00 &  Copyright 2000 Academic Press. &  Copyright 2000 Academic Press. \\
100.00 &  1988). &  (1988) J. \\
100.00 &  Biol. &  Biol. \\
100.00 &  Acad. &  Acad. \\
100.00 &  Virol. &  Virol. \\
100.00 &  1995. &  (1995) J. \\
100.00 &  Natl. &  Natl. \\
100.00 &  Copyright 1999 Academic Press. &  Copyright 1999 Academic Press. \\
100.00 &  The activated glucocorticoid receptor forms a complex with Stat5 and enhances Stat5-mediated transcriptional induction. &  The activated glucocorticoid receptor forms a complex with Stat5 and enhances Stat5-mediated transcriptional induction. \\
100.00 &  Copyright 1999 Academic Press. &  Copyright 1999 Academic Press. \\
100.00 &  Chem. &  Chem. \\
100.00 &  Appl. &  Appl. \\
100.00 &  Copyright 1998 Academic Press. &  Copyright 1998 Academic Press. \\
100.00 &  Sci. &  Sci. \\
100.00 &  (1992) J. &  (1992) J. \\
100.00 &  Acad. &  Acad. \\
100.00 &  Mutational analysis of yeast CEG1 demonstrated that four of the five conserved motifs are essential for capping enzyme function in vivo. &  Mutational analysis of yeast CEG1 demonstrated that four of the five conserved motifs are essential for capping enzyme function in vivo. \\
100.00 &  We also show that in fusions with the DNA binding domain of GAL4, full activity requires the entire BHV-alpha TIF, although both amino and carboxyl termini display some activity on their own. &  We also show that in fusions with the DNA binding domain of GAL4, full activity requires the entire BHV-alpha TIF, although both amino and carboxyl termini display some activity on their own. \\
\hline
\end{tabular}
}
    \caption{Samples with over 75\% similarity in the BC2GM dataset}
    \label{tab:append:bc2gmsim}
\end{table*}

\begin{table*}[h]
    \centering
{\tiny
   \begin{tabular}{p{0.20\linewidth}p{0.10\linewidth}p{0.60\linewidth}}
\hline
Gene & Position & Input \\
\hline
 capping enzyme &  88 100 &  Mutational analysis of yeast CEG1 demonstrated that four of the five conserved motifs are essential for capping enzyme function in vivo. \\
 human IL-4 cytokine gene &  145 165 &  Analysis of 1 Mb of published sequence from the region of conserved synteny on human chromosome 5q31-q33 identified 45 gene candidates, including 35 expressed genes in the human IL-4 cytokine gene cluster. \\
 LH &  1 2 &  (LH P < 0.05, LH/FSH P < 0.01). \\
 LH &  10 11 &  (LH P < 0.05, LH/FSH P < 0.01). \\
 FSH &  13 15 &  (LH P < 0.05, LH/FSH P < 0.01). \\
 Urease &  0 5 &  Urease activity, judged as the amount of ammonia production from urea, could be measured at 25 ng per tube (S/N = 1.5) with Jack bean meal urease. \\
 Jack bean meal urease &  101 118 &  Urease activity, judged as the amount of ammonia production from urea, could be measured at 25 ng per tube (S/N = 1.5) with Jack bean meal urease. \\
 cAMP-CRP &  117 124 &  The reconstituted RNA polymerases containing the mutant alpha subunits were examined for their response to transcription activation by cAMP-CRP and the rrnBP1 UP element. \\
 HIV-2 nef &  51 58 &  These results demonstrate a specific association of SIV and HIV-2 nef, but not HIV-1 nef, with TCRzeta. \\
 HIV-1 nef &  66 73 &  These results demonstrate a specific association of SIV and HIV-2 nef, but not HIV-1 nef, with TCRzeta. \\
\hline
\end{tabular}
}
    \caption{Test samples where the model failed to detect genes, lowering recall, despite the input raw text being an exact match to the training sample}
    \label{tab:append:bc2gmrecallfail}
\end{table*}

\end{document}